\documentclass[lettersize,journal]{IEEEtran}
\usepackage{amsmath,amsfonts}
\usepackage{algorithmic}
\usepackage{algorithm}
\usepackage{array}
\usepackage[caption=false,font=normalsize,labelfont=sf,textfont=sf]{subfig}
\usepackage{textcomp}
\usepackage{stfloats}
\usepackage{url}
\usepackage{verbatim}
\usepackage{graphicx}
\usepackage{cite}
\begin{document}

\title{Distraction is All You Need for Fairness}

\author{Mehdi Yazdani-Jahromi, Amirarsalan Rajabi, Ali Khodabandeh Yalabadi, Aida Tayebi, Ozlem Ozmen Garibay}



\maketitle

\begin{abstract}
Bias in training datasets must be managed for various groups in classification tasks to ensure parity or equal treatment. With the recent growth in artificial intelligence models and their expanding role in automated decision-making, it is vital to ensure that these models are not biased. There is an abundance of evidence suggesting that these models could contain or even amplify the bias present in the data on which they are trained, inherent to their objective function and learning algorithms; Existing methods for mitigating bias result in information loss and do not provide a suitable balance between accuracy and fairness or do not ensure limiting the biases in training. To this end, we propose a powerful strategy for training deep learning models called the Distraction module, which can be effective in controlling bias from affecting the classification results. This method can be utilized with different data types (e.g., tabular, images, graphs). We demonstrate the potency of the proposed method by testing it on \emph{UCI Adult} and \emph{Heritage Health} datasets (tabular), \emph{POKEC-Z}, \emph{POKEC-N} and \emph{NBA} datasets (graph), and \emph{CelebA} dataset (vision). Considering state-of-the-art methods proposed in the fairness literature for each dataset, we exhibit that our model is superior to these proposed methods in minimizing bias and maintaining accuracy.
\end{abstract}

\begin{IEEEkeywords}
Deep Learning, Fairness, Neural Networks, Adversarial Training, Game Theory
\end{IEEEkeywords}

\section{Introduction}
Artificial intelligence and machine learning models in real-world applications have grown in past decades and led to automated decision-making in different domains such as hiring pipelines, face recognition, financial services, the healthcare system, and criminal justice. Algorithmic decision-making may cause an algorithmic bias toward the central population subgroup and discrimination and unfairness toward the minority. In recent years, fairness in artificial intelligence has increased ethical concerns and received attention from interdisciplinary research communities \cite{pessach2020algorithmic}.
Several definitions of fairness have been put forth as potential solutions to the problem of unwanted bias in machine learning techniques. In most cases, the definitions may be separated into two categories: individual fairness and collective fairness. A system that is individual will treat users that are similar to each other in the same manner, where the similarities between people may be determined by past information \cite{dwork2012fairness, yurochkin2019training}. Group fairness metrics are measurements of the statistical equality between different subgroups that are characterized by sensitive characteristics such as race, or gender \cite{zemel2013learning, louizos2015variational, hardt2016equality}.\\
In this paper, our focus is on group fairness, and from this point on, we refer to group fairness as fairness. \\
We introduced a novel in-process bias mitigation method that does not need adversarial example generation to train a fair classifier and can be trained on available datasets without alteration. we proposed a departure from the typical adversarial paradigm. Rather than using a separate discriminator network or injecting adversarial noise, we introduce a new training approach that embeds a specialized module (Distraction Module) within the main neural network. This embedded module independently manages a subset of weights within the main network and is explicitly trained to optimize the network towards specific criteria or alternative objective functions.\\
Our primary contributions are:

\begin{list}{}{}
	\item{A novel training method that incorporates a specialized module within the main network to optimize towards specific criteria or alternative objective functions.}
	\item{Empirical validation of our approach on multiple benchmarks with various data types, demonstrating improvements in network fairness.}
	\item{New adversarial training procedure which significantly improves the state-of-the-art in both accuracy and fairness metrics.}
\end{list}

\section{Related works}
Fairness in machine learning has garnered significant attention, with methods primarily spanning three categories:\\
\textbf{Pre-process approaches} adjust data prior to model training to achieve fair outcomes. This includes changing or reweighing labels \cite{kamiran2012data,luong2011k} and modifying feature distributions to make differentiation between privileged and unprivileged groups \cite{feldman2015certifying, tayebi2022unbiaseddti}. Recently, a GAN was introduced to generate unbiased tabular datasets, focusing on both accuracy and fairness \cite{rajabi2021tabfairgan}. \\
\textbf{In-process approaches} alter the algorithm during training. Some add regularization terms to the objective function, balancing fairness and accuracy. For example, mutual information between protected attributes and classifier predictions was penalized \cite{kamishima2012fairness}, while others added constraints to satisfy equalized odds proxies \cite{zafar2017fairness,zafar2017fairness_a}. \\
\textbf{Post-process approaches} modify outcomes post-training. Strategies range from flipping certain outcomes \cite{hardt2016equality} to using different thresholds for privileged and unprivileged groups, balancing fairness and accuracy \cite{menon2018cost,corbett2017algorithmic}.\\
Different data types, namely tabular, graph, and images, require specialized fairness approaches:\\
\textbf{Tabular}: Efforts in tabular data focus on mitigating bias. Approaches include regularizing covariance between predictions and sensitive variables \cite{cotter2019training}, standardizing decision bounds \cite{zafar2017fairness}, and restricting adversaries from inferring sensitive characteristics \cite{zhang2018mitigating}. Game-theoretic methods have been employed, although scaling them remains challenging \cite{chuang2021fair}. Other noteworthy techniques leverage attention-based approaches, mutual information, and information-theoretic methods \cite{mehrabi2021attributing, moyer2018invariant, song2019learning, jaiswal2020invariant, gupta2021controllable}.\\
\textbf{Graph}: Graphs can amplify biases, particularly in networks where nodes with similar sensitive features are more likely to connect \cite{dong2016young, rahman2019fairwalk}. This can lead to severe decision-making biases in Graph Neural Networks (GNNs) \cite{dai2021say}. Most fair models were designed for i.i.d data and often don't cater to graph data. However, there have been pioneering efforts in learning fair node representations from graphs \cite{bose2019compositional, rahman2019fairwalk, dai2021say}.\\
\textbf{Vision}: Biases in vision models can manifest in various ways, such as gender biases in action recognition or disparities in face recognition across racial and gender categories \cite{zhao2017men,zhao2017men}. Solutions span from altering GAN utility functions for fair image datasets \cite{sattigeri2019fairness, hwang2020fairfacegan} to adversarial game formulations \cite{roy2019mitigating} and methodologies for balanced data generation \cite{ramaswamy2021fair}. Apart from GANs, techniques like U-Nets, deep information maximization adaptation networks, reinforcement learning, and adversarial learning have been proposed to tackle bias in image datasets \cite{rajabi2022through, wang2019racial, wang2019mitigate, chuang2021fair}.
\section{Methodology}

\subsection{Problem Definition}
Our focus in this study is binary classification tasks. We posit that the techniques proposed here can be extended to multi-class datasets without constraint.\\
Let's consider a dataset denoted as \(E = \{x^{(i)}, a^{(i)}, y^{(i)}\}\), where \(x^{(i)}\), \(y^{(i)}\), and \(a^{(i)}\) are independently and identically distributed samples drawn from the data distribution \(P(x, y, a)\). 

Here:
\begin{itemize}
    \item \(x\) represents the features of the dataset.
    \item \(y \in \{0, 1\}\) indicates the label.
    \item \(a\), which takes on discrete finite values, is the protected attribute of our data.
\end{itemize}

\subsection{Model Architecture}
\begin{figure*}
	\vskip 0.2in
	\begin{center}
		\centerline{\includegraphics[width=0.9\textwidth]{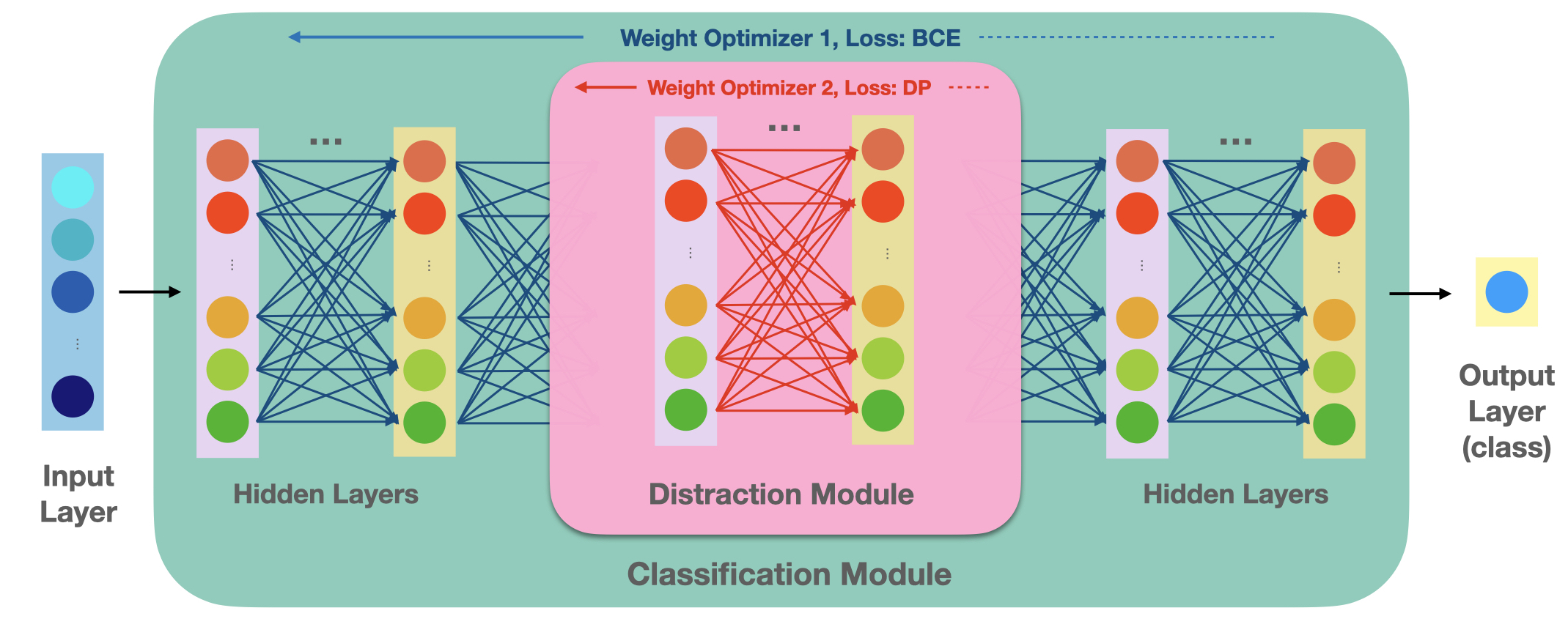}}
		\caption{Depiction of Distraction module in MLP model: The red weights are controlled by the Distraction module and are optimized with Eq. \ref{eq:utility_d} objective function, based on Demographic parity (DP) loss. The blue weights have different optimizer and Binary Cross Entropy (BCE) loss function. They are controlled by the main classifier.}
		\label{fig:model_main}
	\end{center}
	\vskip -0.2in
\end{figure*}

Proposed method employs two distinct sets of weights for its neural network classifier:
\begin{itemize}
    \item The primary set, aimed at the classification task, optimizes for accuracy ($\theta_c$).
    \item The second set, associated with the "Distraction module", is designed to enhance fairness within the model ($\theta_d$).
\end{itemize}
A graphical representation of this architecture can be found in Figure \ref{fig:model_main}. This model can be configured using a variety of neural network architectures such as fully connected layers, graph convolution network, convolution neural network, and etc.

\subsection{Formulation}
The goal of fairness in classification is to ensure that predictions remain consistent regardless of the protected attribute. This concept can be formally captured by:
\begin{equation}
\begin{split}
    \label{eq:invariance}
	&p(C(x, D(x;\theta_{d});\theta_{c}) =  s | a,y) =
	\\& p(C(x, D(x;\theta_{d});\theta_{c}) =  s | a^\prime, y)
 \end{split}
\end{equation}
as described in \cite{louppe2016learning}.

The methodology incorporates a game-theoretic approach. We treat the Distraction module and the encompassing network as two distinct players in a maximin game. Our objective is to train both sets of weights concurrently, aiming to create a classifier that balances fairness with accuracy.
The weights of the Distraction module are entirely isolated from the leading network. The Distraction module tries to make the classifier function results as fair as possible, while the whole network tries to make the classifier function results as accurate as possible. The first player is the Distraction module, and the second is the whole network containing the Distraction module. We train two sets of weights simultaneously to achieve a fair and accurate classifier. This game is established with two utility functions, one for each player. We denote the Distraction module as $D(x)$ and the whole network as $C(x, D(x))$ throughout this paper. To put it differently, D and C play the two-player maximin game with a utility function of $U_1(D, C)$ (Eq. \ref{eq:utility_c}) for player C and $U_2(D, C)$ (Eq. \ref{eq:utility_d}) for player D.
\begin{equation}
	\label{eq:utility_c}
	\min_{\theta_{c}} U_1(C,D)
	 = \mathbb{E}_{x \sim X}\mathbb{E}_{y \sim Y}\Big[- \log p_{\theta_d, \theta_c}(y|x))\Big]
\end{equation}

\begin{equation}
\begin{split}
\label{eq:utility_d}
	\max_{\theta_{d}} U_2(C,D) =
	&-\mathbb{E}_{s \sim C(x, D(x; \theta_d); \theta_c)}\Big[
	\\&\mathbb{E}_{a \sim A}\big[ -\log p_{\theta_d, \theta_c}(a|s)\big]\Big]
 \end{split}
\end{equation}

\subsection{Trade-off Between Fairness and Accuracy}
A common challenge is that a classifier might not be both completely fair and optimal. This is often because the protected attributes, along with their proxies, can influence classifier decisions substantially.

To manage this, we introduce a parameter, \(\eta\), which helps strike a balance between model accuracy and fairness. The role of \(\eta\) is central to the Demographic parity loss (Eq. \ref{eq:eta}) utilized by the Distraction module. A smaller \(\eta\) leans towards accuracy, while a larger value promotes fairness. By tuning \(\eta\), we can generate a series of Pareto solutions for our multi-objective optimization task.
\begin{equation}
    \label{eq:eta}
    \eta\sum_{i=1}^{m}  -\log p_{\theta_d, \theta_c}(a^{(i)}|C(x^{(i)},D(x^{(i)})))
\end{equation}

\subsection{Training Procedure}
We employ a mini-batch stochastic gradient descent technique to train our network. Algorithm \ref{alg:training} offers a detailed training procedure used for training networks containing Distraction module.\\ 

During each iteration:
\begin{itemize}
    \item First, the network's fairness loss is computed, followed by an update to the Distraction module weights.
    \item Then, the network undergoes another iteration with frozen weights of the Distraction module where the classification loss is computed, and back-propagation is employed for updating the weight of the remaining network.
\end{itemize}

\begin{algorithm}
	\caption{Minibatch stochastic gradient descent for adversarial training of a network with distraction module}
	\label{alg:training}
	\begin{algorithmic}
		\STATE {\bfseries Input:} data $(X, A, Y)$, A is a set of protected attributes and Y is the label, Batch Size $m$, $C$-Learning rate $lr_1$, $D$-Learning rate $lr_2$, $\eta$ 
		\FOR{number of iterations in training}
		\STATE sample minibatch of size $m$ samples $\{x^{(1)}, x^{(2)}, ... , x^{(m)}\}$ from data $P(x)$
		\STATE updates the Distraction module by ascending the stochastic gradient with learning rate $lr_2$:
		\begin{eqnarray}
			\nonumber
			\nabla_{\theta_{d}}\eta\sum_{i=1}^{m}  -\log p_{\theta_d, \theta_c}(a^{(i)}|C(x^{(i)},D(x^{(i)})))
		\end{eqnarray}
		\STATE updates the classifier network by descending the stochastic gradient  with learning rate $lr_1$:
		\begin{eqnarray}
			\nonumber
			\nabla_{\theta_{c}} \sum_{i=1}^{m} -\log p_{\theta_d, \theta_c}(y^{(i)}|x^{(i)})
		\end{eqnarray}
		\ENDFOR
		\STATE In a practical perspective, this method needs two optimizers which can be any standard gradient-based method. We used the Adam optimizer for both of the functions in our experiments.
	\end{algorithmic}
\end{algorithm}
\section{Experiments}
\begin{table}[t]
	\caption{Area over the curve of statistical demographic parity and accuracy (Higher is better). The proposed model (Distraction) significantly outperforms other benchmark models in this quantitative metric.}
	\label{tab:area_over_the_curve}
	\begin{center}
				\begin{tabular}{l |c c}
                    \hline
					Method & UCI Adult & Heritage Health \\
                    \hline
					Distraction (Ours)  & \textbf{0.411} & \textbf{0.503} \\
					FCRL {\tiny(AAAI 2021)} & 0.253 & 0.285 \\
					Attention & 0.213 & 0.139 \\
					CIVB {\tiny(NeurIPS 2018)} & 0.163 & 0.176 \\
					MIFR {\tiny(AIStats 2019)} & 0.221 & 0.166 \\
					MaxEnt-ARL {\tiny(CVPR 2019)} & 0.133 & 0 \\
					Adv Forgetting {\tiny(AAAI 2020)} & 0.077 & 0.172 \\
                \hline
				\end{tabular}
	\end{center}
\end{table}
This section compares our method to other benchmark methods in the literature. The experiment section consists of three parts. First, we experiment with tabular datasets. We compare the classification accuracy and statistical parity of deep learning methods in the benchmark datasets. In the second section, we use graph data for node classification tasks. We evaluate our model on vision datasets in the third and final section.\\
The Distraction module used on all the datasets consists of only linear layers, and the Distraction module is positioned one layer before the classification layer. This choice was due to experiments conducted on the vision and tabular datasets. The ablation study, and the additional results can be found in the section \ref{ap:ablation}.
Additionally, we observed that the loss for both the fairness metric and the accuracy is very volatile in training. However, given enough steps, it always converges to a single point which is a Pareto answer for this multi-objective optimization. We decided not to include loss graphs in the paper due to the volatility of the losses. 
\subsection{Evaluation Metrics}
\label{ap:metric}
We employ four evaluation metrics to compare the performance of our model with baseline models. These metrics are as follows:

\begin{enumerate}
    \item \textbf{Average Precision (AP)}:
    \begin{itemize}
        \item \textit{Definition}: The average precision is a measure that computes the average while combining recall and precision at each threshold. It provides an aggregate assessment of the classifier's performance over all levels of precision-recall.
        \item \textit{Objective}: Higher AP values are preferred, indicating better accuracy of the classifiers.
    \end{itemize}

    \item \textbf{Accuracy}:
    \begin{itemize}
        \item \textit{Usage}: Consistency with literature suggests using accuracy for tabular and graph datasets.
        \item \textit{Objective}: A higher accuracy indicates a better performing model.
    \end{itemize}

    \item \textbf{Demographic Parity (DP)}:
    \begin{itemize}
        \item \textit{Definition}: DP is a widely-used fairness metric that captures the difference in the probability of receiving a favorable decision between different protected groups. It is calculated as the absolute difference $(|P(Y = 1|S = 0) - P(Y = 1|S = 1)|)$. For more than two groups, DP can be computed using $\Delta_{DP} (a, \hat{y}) = \max_{a_i, a_j} |P(\hat{y} = 1 | a=a_i) - P(\hat{y} =1 | a=a_j))|$~\cite{mehrabi2021survey,gupta2021controllable}.
        \item \textit{Objective}: A smaller DP indicates a more fair classification, as it reduces disparity between protected groups.
    \end{itemize}

    \item \textbf{Difference in Equality of Opportunity ($\Delta$EO)}:
    \begin{itemize}
        \item \textit{Definition}: Following~\cite{lokhande2020fairalm} and~\cite{ramaswamy2021fair}, we use $\Delta$EO as a fairness metric. It is defined as the absolute difference between the true positive rates for different group expressions $(|TPR(S = 0) - TPR(S = 1)|)$.
        \item \textit{Objective}: Lower values of $\Delta$EO are preferred, indicating fairer categorization between group expressions.
    \end{itemize}
\end{enumerate}

In the experiments, we used demographic parity as the fairness criterion during training. This choice indicates that the model is optimized for demographic parity and the demographic parity is the main metric that our model is providing. The better fairness metrics provided are byproduct of optimizing the model with the demographic parity objective.

\subsection{Implementation details}
The hyperparameters used in training the models on each tabular, graph, and vision datasets can be found in the tables \ref{hype1}, \ref{hype2}, and \ref{hype3} respectively.\\
The training was performed on a single NVIDIA GeForce RTX 3090.

\begin{table}
\caption{Summary of Parameter Setting for the Distraction on tabular datasets} 
\centering 
\begin{tabular}{l c c } 
\hline
Hyperparameters & UCI Adult & Health Heritage \\
\hline
FC layers before the Distraction module & 2 & 2 \\
FC layers of the Distraction module & 1 & 3\\
FC layers after the Distraction module & 1 & 1 \\
Epoch & 50 & 50 \\
Batch size & 100 & 100 \\
Dropout & 0 & 0 \\
Network optimizer & Adam & Adam\\
Distraction module optimizer & Adam & Adam\\
Network learning rate & 1e-3 & 1e-3 \\ 
Distraction module learning rate & 1e-5 & 1e-5\\
$\eta$ & 100 & 100\\

\hline
\end{tabular}

\label{hype1} 
\end{table}

\begin{table*}
\caption{Summary of Parameter Setting for the Distraction on graph datasets} 
\centering 
\begin{tabular}{l c c c } 
\hline 
Hyperparameters & POKEC-Z & POKEC-N & NBA \\ 
\hline 
GCN layer before the Distraction module & 2 & 2 & 2 \\
Distraction module FC layers & 1 & 1 & 1 \\
FC layers after the Distraction module & 1 & 1 & 1\\
Epoch & 5000 & 1000 & 1000 \\
Batch size & 1 & 1 & 1 \\
Dropout & 0 & 0.5 & 0.5 \\
Network optimizer & Adam & Adam & Adam\\
Distraction module optimizer & Adam & Adam & Adam\\
Network learning rate & 1e-3 & 1e-3 & 1e-2\\ 
Distraction module learning rate & 1e-6 & 1e-8 & 1e-5\\
$\eta$ & 1000 & 100 & 1000\\
\hline
\end{tabular}

\label{hype2} 
\end{table*}

\begin{table*}
\caption{Summary of Parameter Setting for the Distraction on vision dataset} 
\centering 
\begin{tabular}{l c c c } 
\hline 
Hyperparameters & CelebA-Attractive & CelebA-Smiling & CelebA-WavyHair \\ 
\hline 
Distraction module FC layers & 1 & 1 & 1 \\
FC layers after the Distraction module & 1 & 1 & 1\\
Epoch & 30 & 15 & 15 \\
Batch size & 128 & 128 & 128 \\
Dropout & 0 & 0 & 0 \\
Network optimizer & Adam & Adam & Adam\\
Distraction module optimizer & Adam & Adam & Adam\\
Network learning rate & 1e-3 & 1e-3 & 1e-3\\ 
Distraction module learning rate & 1e-6 & 1e-5 & 1e-5\\
$\eta$ & 1000 & 100 & 100\\ 
\hline
\end{tabular}

\label{hype3} 
\end{table*}

\subsection{Tabular}
We evaluated our method for bias mitigation to various current state-of-the-art approaches. We concentrate on strategies specifically tuned to achieve the best results in statistical parity metrics on tabular studies.\\
\subsubsection{Datasets}
We conducted experiments using two well-established benchmark datasets in this field:\\
\textbf{UCI Adult Dataset}~\cite{uciadult}: This dataset, based on demographic data collected in 1994, includes a training set of 30,000 samples and a test set of 15,000 samples. The task is to predict whether an individual's salary exceeds \$50,000 per annum, with gender serving as the binary protected attribute.

\textbf{Heritage Health Dataset}: This dataset involves predicting the Charleson Index, a measure of a patient's 10-year mortality risk. It comprises samples from approximately 51,000 patients, split into a training set of 41,000 and a test set of 11,000. The protected attribute, age, has nine possible values.

We selected the protected and target attributes in accordance with existing literature.

\begin{figure*}[!t]
\centering
\subfloat[]{\includegraphics[width=3.5in]{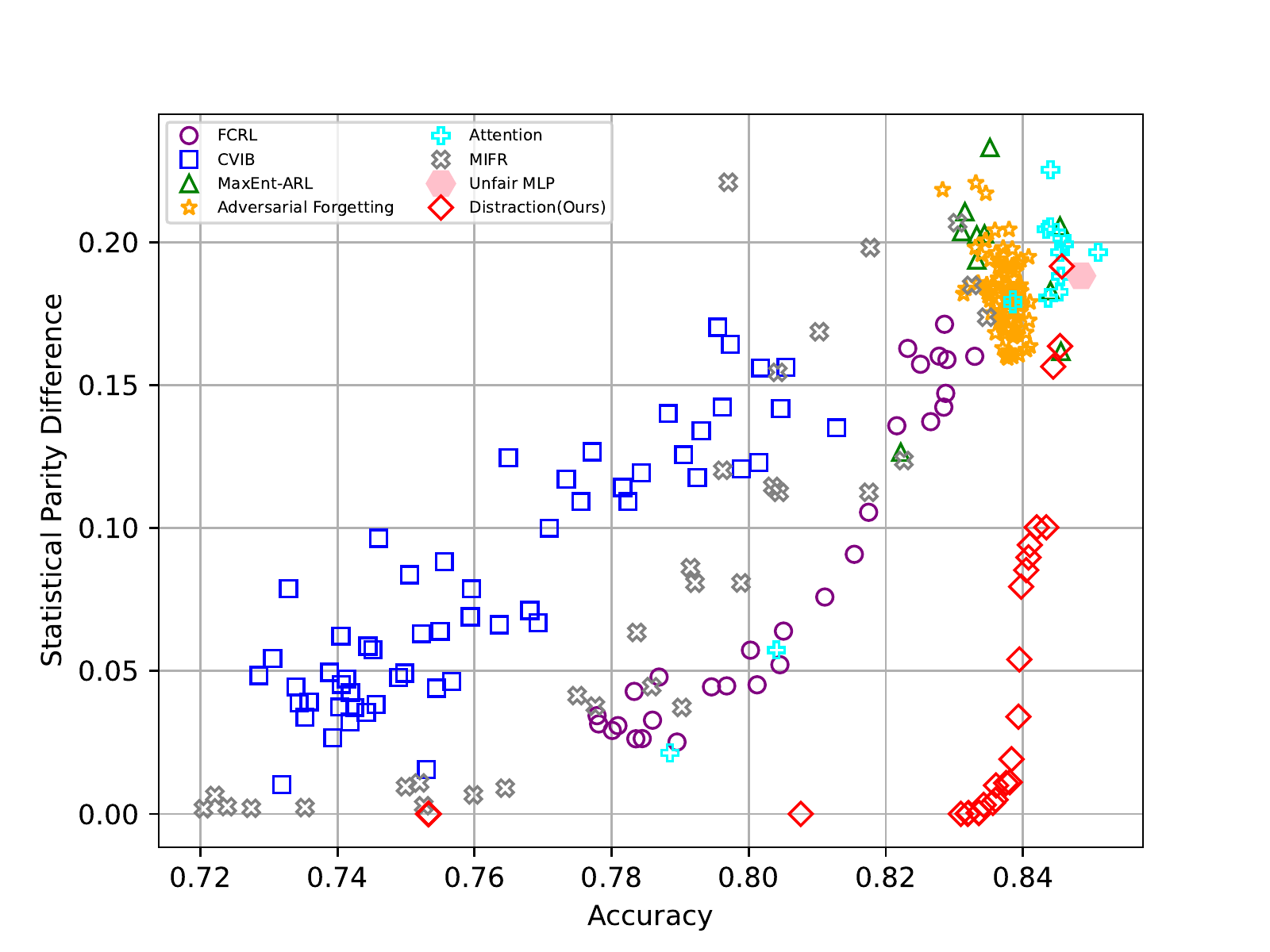}%
\label{fig:adult_compare}}
\hfil
\subfloat[]{\includegraphics[width=3.5in]{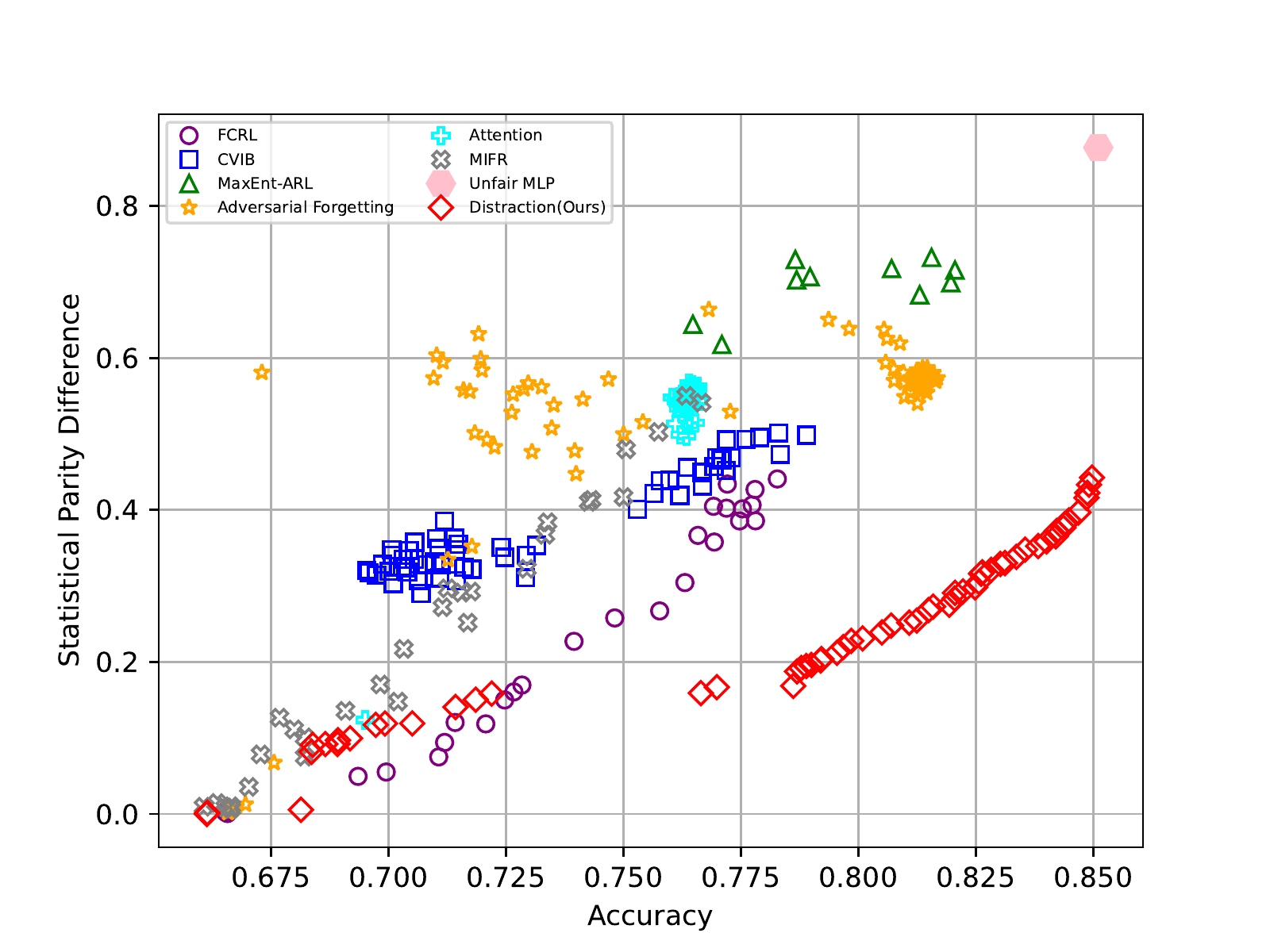}%
\label{fig:health_compare}}
\caption{Accuracy of different benchmark models and distraction model vs. statistical demographic parity of each model for the (a) UCI Adult dataset and (b)  Heritage Health dataset. The ideal area for this graph is the bottom right which indicates high accuracy and low statistical demographic parity. This graph shows that our model performs significantly better than the other benchmark models in both tabular datasets.}
\label{fig_sim}
\end{figure*}



We used training sets to run the Distraction module technique and train the network. Then we assess the performance by running classifiers for subsequent prediction tasks.
\subsubsection{Benchmark Methods}
We compared our approach with the following state-of-the-art techniques:

\begin{itemize}
    \item \textbf{CVIB}~\cite{moyer2018invariant}: This method employs a conditional variational autoencoder for bias mitigation.
    \item \textbf{MIFR}~\cite{song2019learning}: This technique optimizes fairness by leveraging an information bottleneck factor combined with adversarial learning.
    \item \textbf{FCRL}~\cite{gupta2021controllable}: This approach uses specific approximations for contrastive information to maximize theoretical goals, which can be employed to strike appropriate trade-offs between statistical demographic parity and accuracy.
    \item \textbf{MaxEnt-ARL}~\cite{roy2019mitigating}: This method uses adversarial learning to reduce bias in tabular data.
    \item \textbf{Adversarial Forgetting}~\cite{jaiswal2020invariant}: Another method that employs adversarial learning for bias mitigation.
\end{itemize}
\subsubsection{Results}
We employed the training sets to execute our Distraction module method and train the network. Subsequent prediction tasks were evaluated using classifiers. The average accuracy, representing the most likely accuracy, and maximum demographic parity, indicating worst-case bias, were computed over five training iterations with random seeds. In contrast to~\cite{gupta2021controllable}, no preprocessing was applied to the data before inputting it into our network. Our reported results represent Pareto solutions for the neural network during training with varying $\eta$s.

Figures~\ref{fig:adult_compare} and~\ref{fig:health_compare} depict trade-offs between statistical demographic parity and accuracy for various bias reduction techniques on the UCI Adult and Heritage Health datasets, respectively. An ideal result would position the curve in the lower right corner of the graph, indicating accurate and fair outcomes concerning protected attributes. Our results demonstrate that the Distraction method significantly outperforms competing methods. This finding is further supported by the area-over-the-curve data for demographic parity and accuracy (Table~\ref{tab:area_over_the_curve}), showing that our proposed strategy improves the area over the curve for both datasets by a factor of $\sim$2. This implies that our bias reduction framework is the most effective mitigation strategy for tabular data.

\subsection{Graph}
\begin{table*}[ht]
\caption{The comparisons of our proposed method with the baselines on Pokec-z}
\label{tab:pokz}
\vskip 0.15in
\begin{center}
\begin{small}
\begin{sc}
\begin{tabular}{lcccc}
\hline
Method                 & ACC(\%)                  & AUC(\%)                  & $\Delta_{DP}$(\%)                  & $\Delta_{EO}$(\%)         \\
\hline

ALFR                   & 65.4 ±0.3            & 71.3 ±0.3            & 2.8 ±0.5            & 1.1 ±0.4   \\
ALFR-e                 & 68.0 ±0.6            & 74.0 ±0.7            & 5.8 ±0.4            & 2.8 ±0.8   \\
Debias                 & 65.2 ±0.7            & 71.4 ±0.6            & 1.9 ±0.6            & 1.9 ±0.4   \\
Debias-e               & 67.5 ±0.7            & 74.2 ±0.7            & 4.7 ±1.0            & 3.0 ±1.4   \\
FCGE                   & 65.9 ±0.2            & 71.0 ±0.2            & 3.1 ±0.5            & 1.7 ±0.6   \\
FairGCN                & 70.0 ±0.3            & 76.7 ±0.2            & 0.9 ±0.5            & 1.7 ±0.2   \\
FairGAT                & 70.1 ±0.1            & 76.5 ±0.2            & \textbf{0.5 ±0.3}            & 0.8 ±0.3   \\
NT-FairGNN             & 70.0 ±0.1            & 76.7 ±0.3            & 1.0 ±0.4            & 1.6 ±0.2   \\
GAT+Distraction (Ours) & \textbf{70.97 ±0.16} & \textbf{77.58 ±0.13} & 0.93 ±0.44 & \textbf{0.97 ±0.40}\\
\hline
\end{tabular}
\end{sc}
\end{small}
\end{center}
\end{table*}

\begin{table*}[t]
\caption{The comparisons of our proposed method with the baselines on Pokec-n}
\label{tab:pokn}
\vskip 0.15in
\begin{center}
\begin{small}
\begin{sc}
\begin{tabular}{lcccc}
\hline
Method                 & ACC(\%)                  & AUC(\%)                  & $\Delta_{DP}$(\%)                  & $\Delta_{EO}$(\%)         \\
\hline
ALFR                   & 63.1 ±0.6           & 67.7 ±0.5           & 3.05 ±0.5           & 3.9 ±0.6          \\
ALFR-e                 & 66.2 ±0.5           & 71.9 ±0.3           & 4.1 ±0.5            & 4.6 ±1.6          \\
Debias                 & 62.6 ±0.9           & 67.9 ±0.7           & 2.4 ±0.7            & 2.6 ±1.0          \\
Debias-e               & 65.6 ±0.8           & 71.7 ±0.7           & 3.6 ±0.2            & 4.4 ±1.2          \\
FCGE                   & 64.8 ±0.5           & 69.5 ±0.4           & 4.1 ±0.8            & 5.5 ±0.9          \\
FairGCN                & 70.1 ±0.2           & 74.9 ±0.4           & 0.8 ±0.2            & 1.1 ±0.5          \\
FairGAT                & 70.0 ±0.2           & 74.9 ±0.4           & 0.6 ±0.3            & \textbf{0.8 ±0.2} \\
NT-FairGNN             & 70.1 ±0.2           & 74.9 ±0.4           & 0.8 ±0.2            & 1.1 ±0.3          \\
GAT+Distraction (Ours) & \textbf{70.07 ±0.5} & \textbf{75.8 ±0.38} & \textbf{0.62 ±0.14} & 3.0 ±1.0         \\
\hline
\end{tabular}
\end{sc}
\end{small}
\end{center}
\end{table*}

\begin{table*}[t]
\caption{The comparisons of our proposed method with the baselines on NBA}
\label{tab:nba}
\vskip 0.15in
\begin{center}
\begin{small}
\begin{sc}
\begin{tabular}{lcccc}
\hline
Method                 & ACC(\%)                  & AUC(\%)                  & $\Delta_{DP}$(\%)                  & $\Delta_{EO}$(\%)         \\
\hline
ALFR                   & 64.3 ±1.3            & 71.5 ±0.3            & 2.3 ±0.9            & 3.2 ±1.5          \\
ALFR-e                 & 66.0 ±0.4            & 72.9 ±1.0            & 4.7 ±1.8            & 4.7 ±1.7          \\
Debias                 & 63.1 ±1.1            & 71.3 ±0.7            & 2.5 ±1.5            & 3.1 ±1.9          \\
Debias-e               & 65.6 ±2.4            & 72.9 ±1.2            & 5.3 ±0.9            & 3.1 ±1.3          \\
FCGE                   & 66.0 ±1.5            & 73.6 ±1.5            & 2.9 ±1.0            & 3.0 ±1.2          \\
FairGCN                & 71.1 ±1.0            & 77.0 ±0.3            & 1.0 ±0.5            & 1.2 ±0.4          \\
FairGAT                & 71.5 ±0.8            & 77.5 ±0.7            & 0.7 ±0.5            & \textbf{0.7 ±0.3} \\
NT-FairGNN             & 71.1 ±1.0            & 77.0 ±0.3            & 1.0 ±0.5            & 1.2 ±0.4          \\
GAT+Distraction (Ours) & \textbf{77.09 ±0.45} & \textbf{77.99 ±0.58} & \textbf{0.34 ±0.21} & 12.78 ±2.9 \\
\hline
\end{tabular}
\end{sc}
\end{small}
\end{center}
\end{table*}
We compare our proposed framework with state-of-the-art approaches for fair classification and fair graph embedding learning, including ALFR~\cite{edwards2015censoring}, ALFR-e, Debias~\cite{zhang2018mitigating}, Debias-e, FCGE~\cite{bose2019compositional}, FairGCN~\cite{dai2021say}, and NT-FAIRGNN~\cite{dai2023fairgraph}. A brief overview of these methods is as follows:

\begin{itemize}
    \item \textbf{ALFR}~\cite{edwards2015censoring}: A pre-processing approach that removes sensitive information from representations generated by an MLP-based autoencoder using a discriminator. The debiased representations are then used to train a linear classifier.
    \item \textbf{ALFR-e}: An extension of ALFR that incorporates graph structure information by combining user features in ALFR with graph embeddings discovered by deepwalk~\cite{perozzi2014deepwalk}.
    \item \textbf{Debias}~\cite{zhang2018mitigating}: An in-processing fair classification method that directly applies a discriminator to the predicted probability of the classifier.
    \item \textbf{Debias-e}: An extension of Debias that includes deepwalk embeddings into the Debias features.
    \item \textbf{FCGE}~\cite{bose2019compositional}: A method for learning fair node embeddings in graphs without node characteristics. Discriminators filter out sensitive data in the embeddings.
    \item \textbf{FairGCN}~\cite{dai2021say}: A graph convolutional network designed for fairness in graph-based learning. It incorporates fairness constraints during training to reduce disparities between protected groups.
    \item \textbf{NT-FAIRGNN}~\cite{dai2023fairgraph}: A graph neural network that aims to achieve fairness by balancing the trade-off between accuracy and fairness. It uses a two-player minimax game between the predictor and the adversary, where the adversary aims to maximize the unfairness.
\end{itemize}

\subsubsection{Datasets}
We conducted experiments using the following datasets obtained from the study by~\cite{dai2021say}:

\begin{itemize}
    \item \textbf{Pokec}~\cite{takac2012data}: A widely-used social network dataset from Slovakia, akin to Facebook and Twitter, containing anonymized data from the entire social network of  year 2012. The dataset includes user information such as gender, age, interests, hobbies, and profession. Sampled subsets, Pokec-z and Pokec-n, are created based on user provinces. The classification task involves predicting users' working environments.
    \item \textbf{NBA}: A Kaggle dataset of approximately 400 NBA basketball players, featuring 2016-2017 season statistics, nationality, age, and salary. Graph connections were established using relationships between NBA players on Twitter, collected via the official API. Players are categorized as American or international, a sensitive attribute. The classification task is to predict whether a player's wage is above the median.
\end{itemize}

\subsubsection{Results}
Each experiment was conducted five times, and Tables~\ref{tab:pokz},~\ref{tab:pokn}, and~\ref{tab:nba} report the mean and standard deviation of the runs for Pokec-z, Pokec-n, and NBA datasets, respectively. These results represent the selected Pareto solutions for comparison with the benchmarks. The tables reveal that, in comparison to GAT, generic fair classification techniques and graph embedding learning approaches exhibit inferior classification performance, even when utilizing graph information. In contrast, our Distraction method performs comparably to baseline GNNs. FairGCN is close to the baseline, but our Distraction technique outperforms it. When sensitive information is scarce (e.g., NBA dataset), baselines exhibit clear bias, with graph-based baselines performing worse. However, our proposed model yields near-zero statistical demographic parity, indicating effective discrimination mitigation.

\subsection{Vision}

\begin{figure*}[!t]
\centering
\subfloat[]{\includegraphics[width=3.5in]{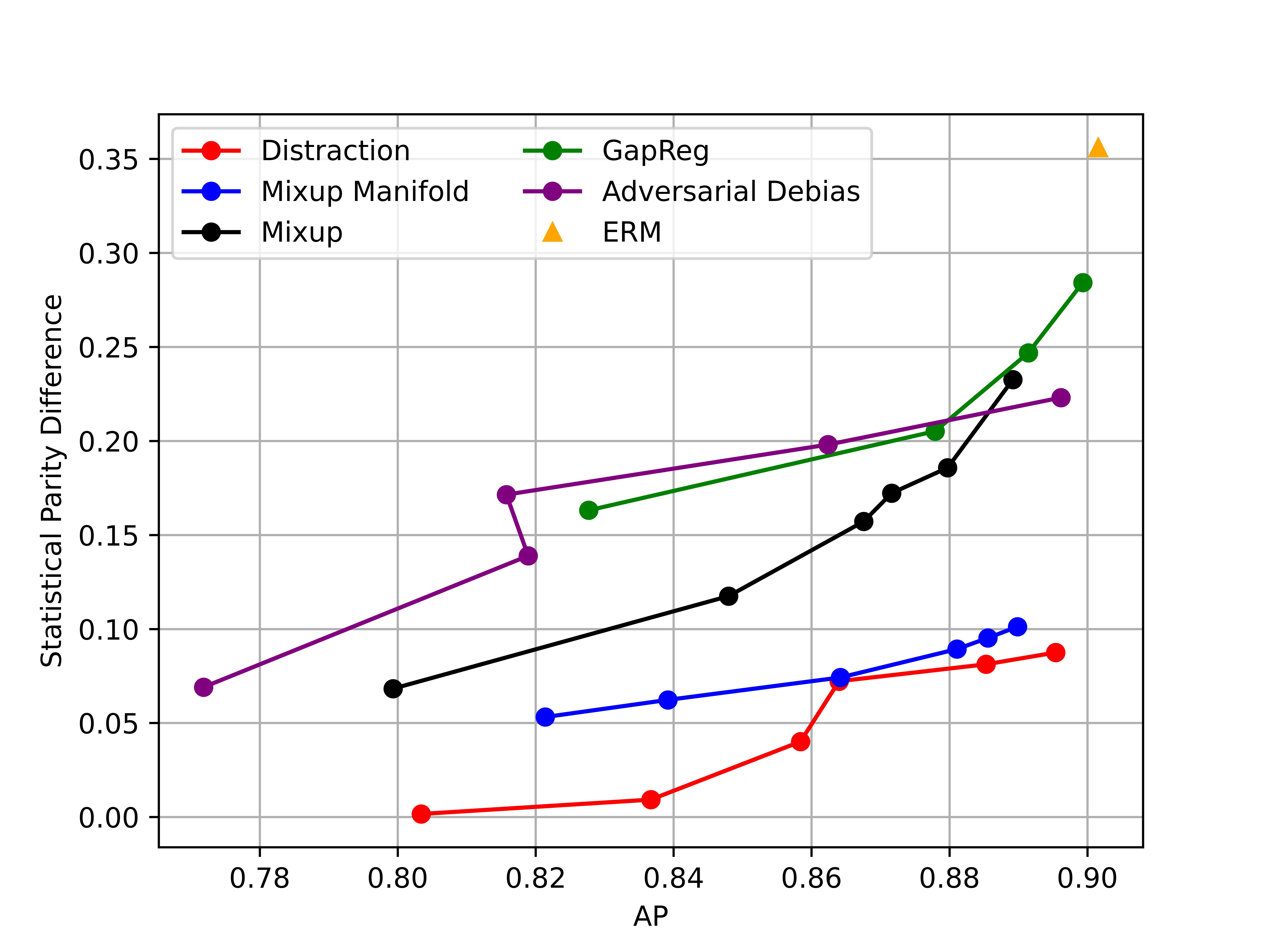}%
\label{fig:1}}
\hfil
\subfloat[]{\includegraphics[width=3.5in]{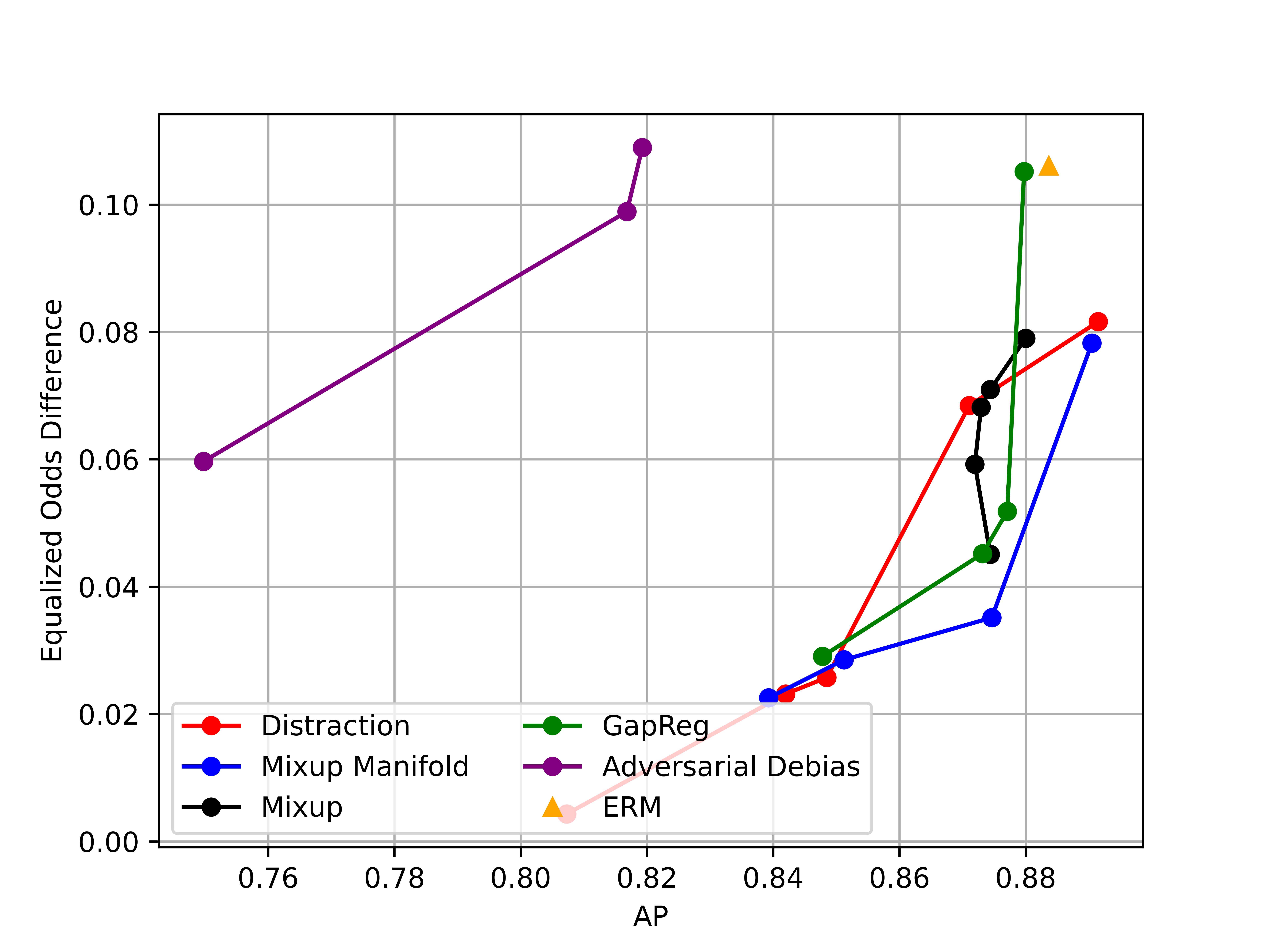}%
\label{fig:4}}
\caption{"Attractive" Attribute of CelebA Dataset as the Target Attribute. (a) reflects the trade-off between Average Precision and Demographic Parity Difference. (b) shows the trade-off between Average Precision and Equalized Odds Difference. the Distraction module is showing competitive results to the baseline.}
\label{fig:celebA_attractive}
\end{figure*}

\begin{figure*}[!t]
\centering
\subfloat[]{\includegraphics[width=3.5in]{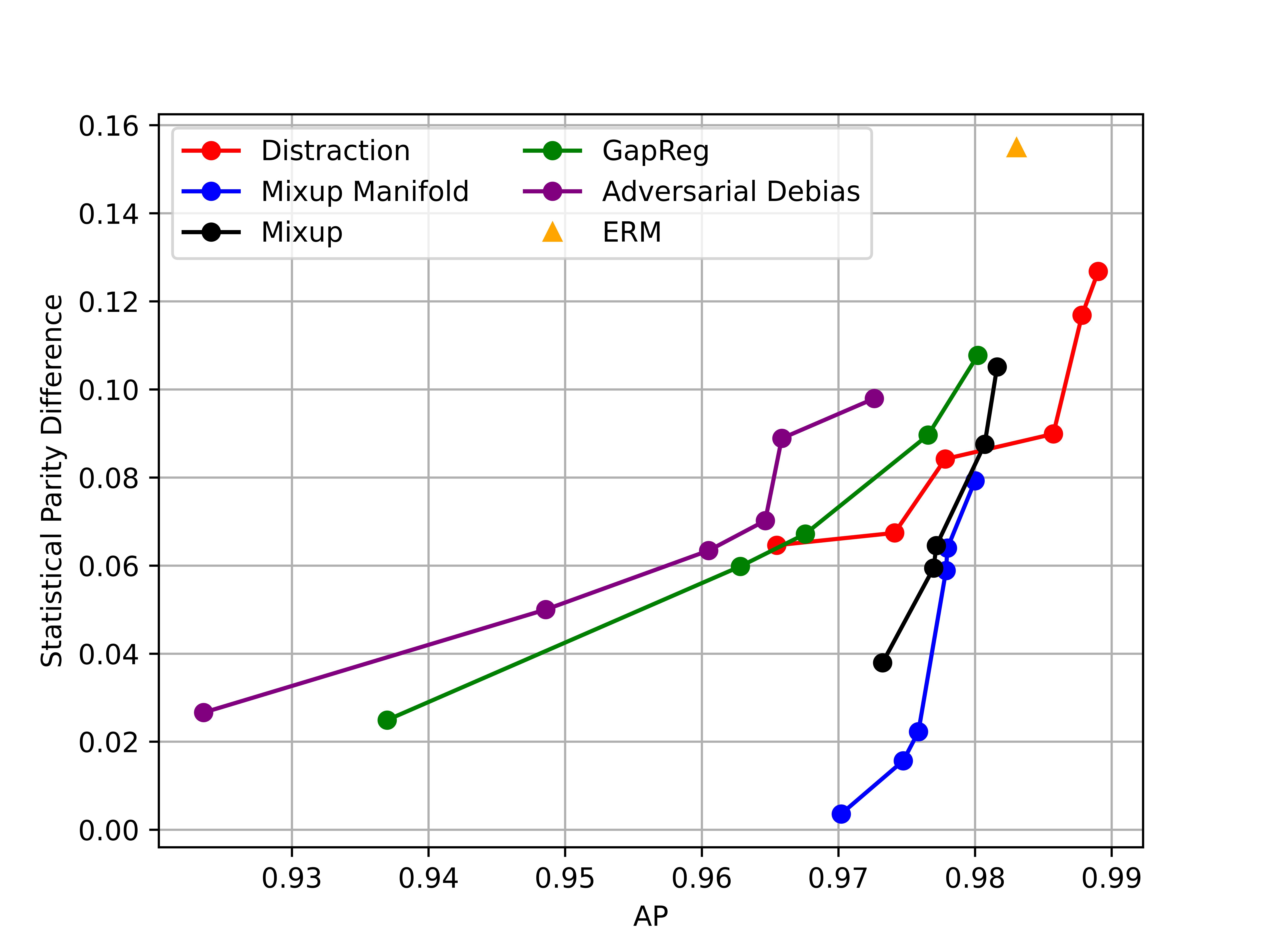}%
\label{fig:2}}
\hfil
\subfloat[]{\includegraphics[width=3.5in]{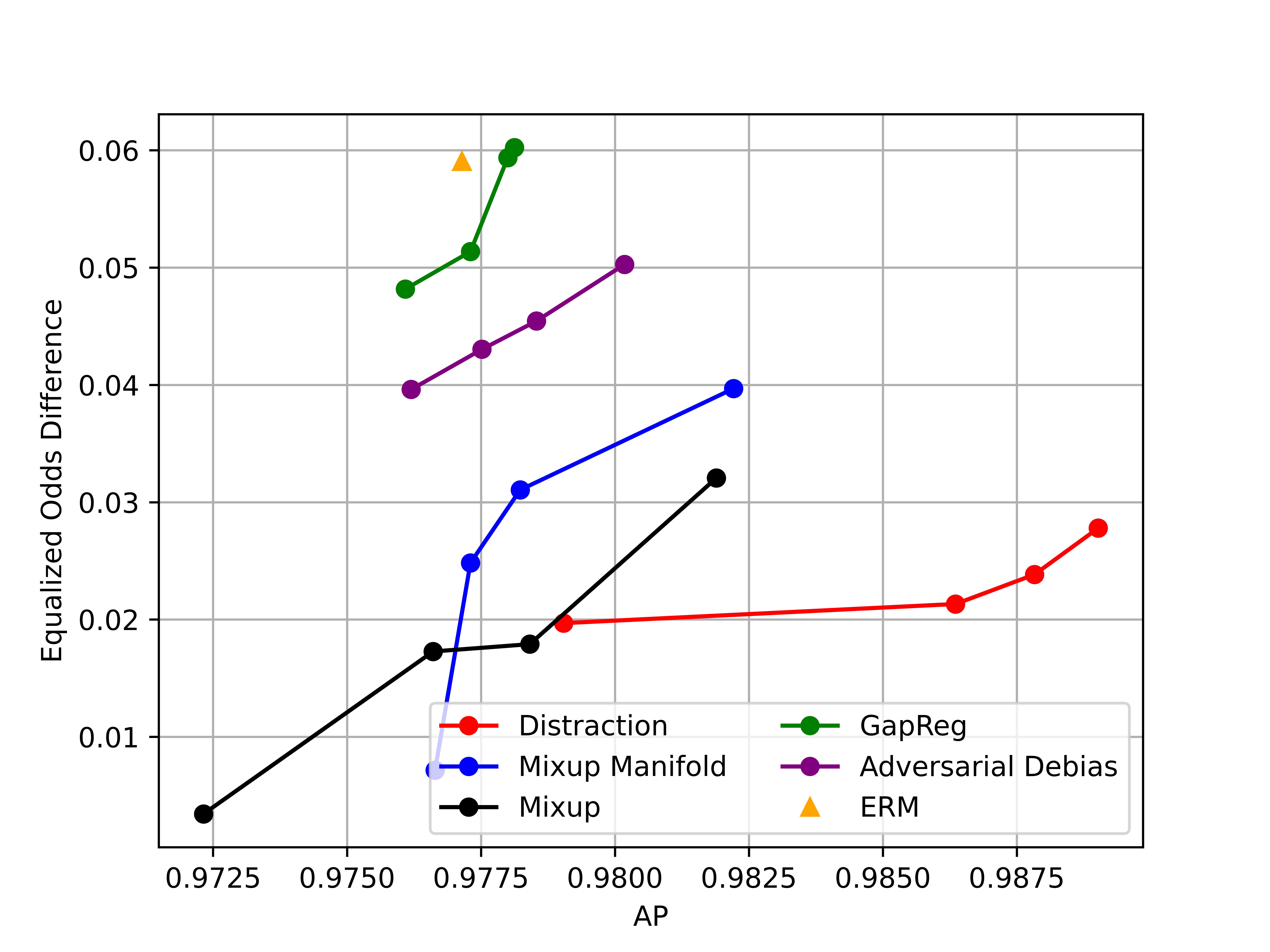}%
\label{fig:5}}
\caption{\emph{"Smiling" Attribute of CelebA Dataset as the Target Attribute}. (a) reflects the trade-off between Average Precision and Demographic Parity Difference. (b) shows the trade-off between Average Precision and Equalized Odds Difference. the Distraction module is showing competitive results to the baseline.}
\label{fig:celebA_smiling}
\end{figure*}

\begin{figure*}[!t]
\centering
\subfloat[]{\includegraphics[width=3.5in]{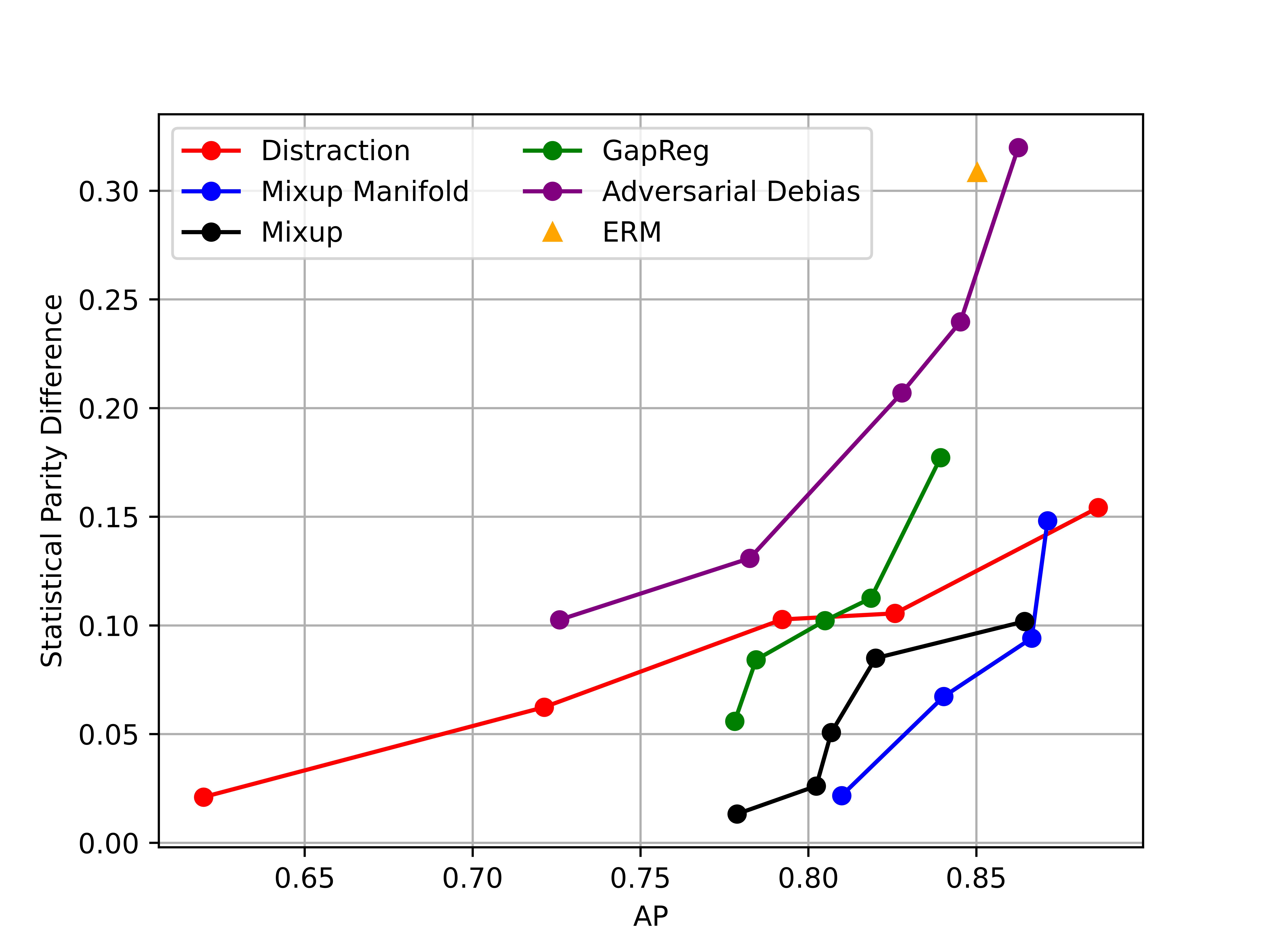}%
\label{fig:3}}
\hfil
\subfloat[]{\includegraphics[width=3.5in]{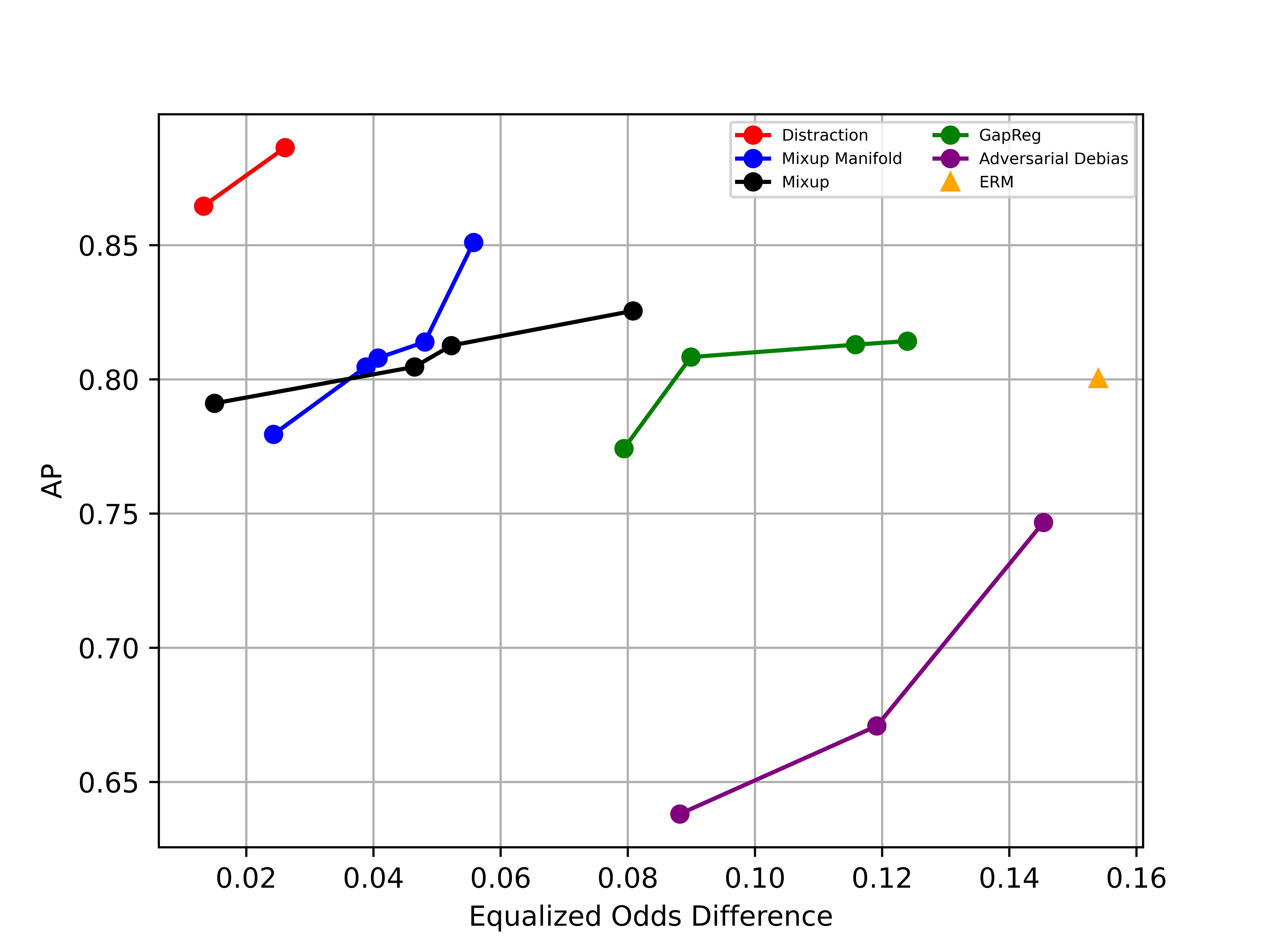}%
\label{fig:6}}
\caption{\emph{"Wavy Hair" Attribute of CelebA Dataset as the Target Attribute}. (a) reflects the trade-off between Average Precision and Demographic Parity Difference. (b) shows the trade-off between Average Precision and Equalized Odds Difference. the Distraction module is showing competitive results to the baseline.}
\label{fig:celebA_wavy}
\end{figure*}

We compare our proposed method on vision tasks with several existing approaches, including: (1) Empirical Risk Minimization (ERM) which achieves the training task without any regularization, (2) Gap Regularization which directly regularizes the model, (3) Adversarial Debiasing~\cite{zhang2018mitigating}, and (4) FairMixup~\cite{chuang2021fair}. 

\subsubsection{Dataset and Setup}
To demonstrate the efficacy of our method, we used the CelebA dataset of face attributes~\cite{liu2015faceattributes}, consisting of over 200,000 images of celebrities. Each image in this dataset has been annotated with 40 binary attributes, including gender, by human annotators. We chose three attributes -- attractive, smiling, and wavy hair -- for binary classification tasks, using gender as the protected attribute. These attributes were selected because each of them has a sensitive group that receives a disproportionately high number of positive samples. For each task, we employed a ResNet-18 architecture \cite{resnet}, augmented with two additional layers for outcome prediction.

\subsubsection{Results}
The trade-off between Average Precision (AP), Demographic Parity (DP), and  Equality of Opportunity (EO) for attributes "Attractive", "Smiling", and "Wavy Hair" is illustrated in the figures \ref{fig:celebA_attractive}, \ref{fig:celebA_smiling}, and \ref{fig:celebA_wavy} respectively. 
Our proposed Distraction module provides a more balanced trade-off between accuracy and fairness. Instead of prioritizing one over the other, our method strikes a better balance, ensuring that the trained model is both accurate and fair.
Moreover, our Distraction module consistently provides better equality of opportunity across various accuracy levels compared to benchmark models. Through empirical validation on multiple benchmarks, we've shown that the Distraction module consistently outperforms other methods in achieving equality of opportunity across various accuracy levels. This indicates that our method can provide fair treatment to different protected groups while still maintaining high predictive accuracy.
\begin{figure*}[!t]
\centering
\subfloat[]{\includegraphics[width=3.5in]{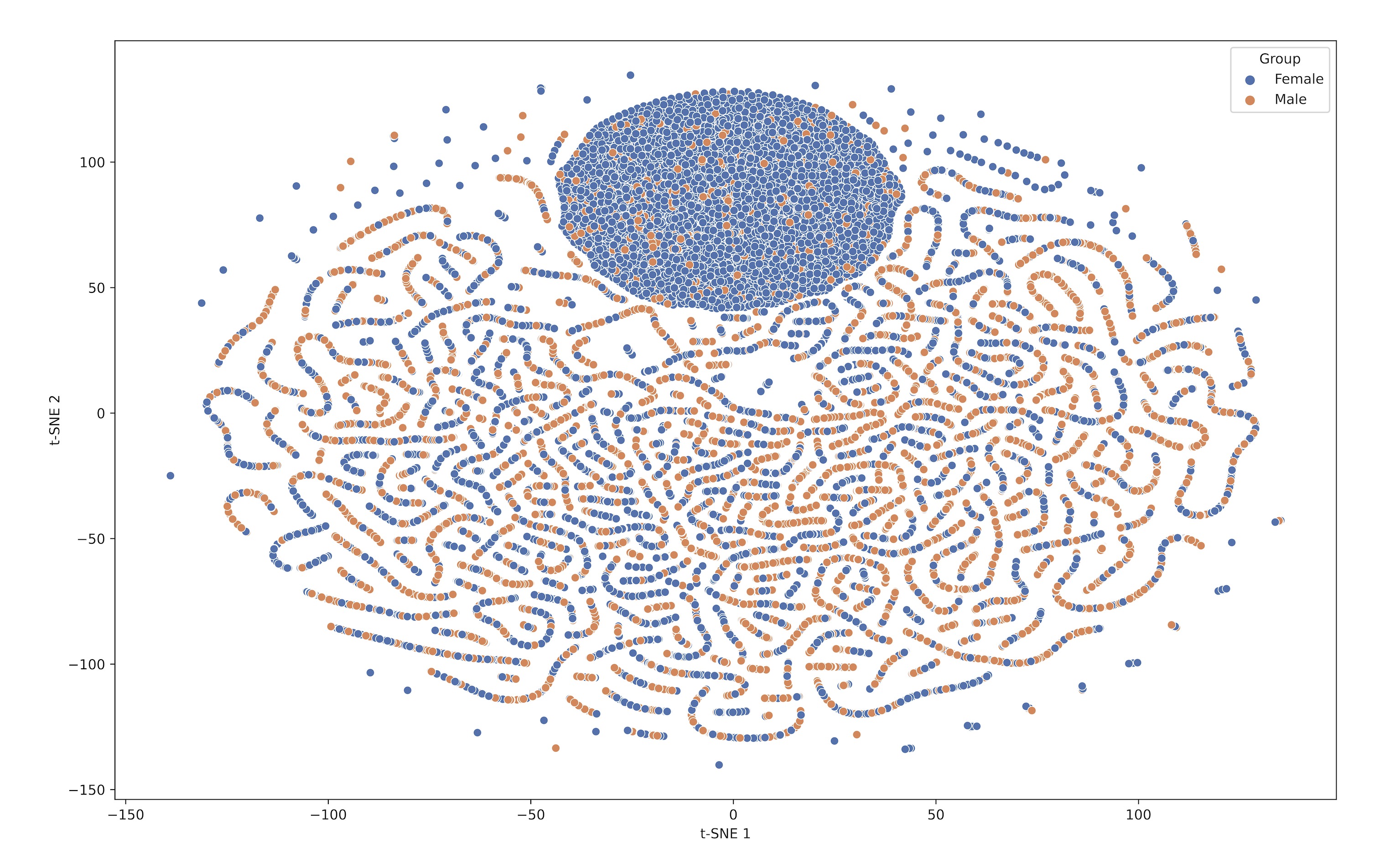}%
\label{fig:4}}
\hfil
\subfloat[]{\includegraphics[width=3.5in]{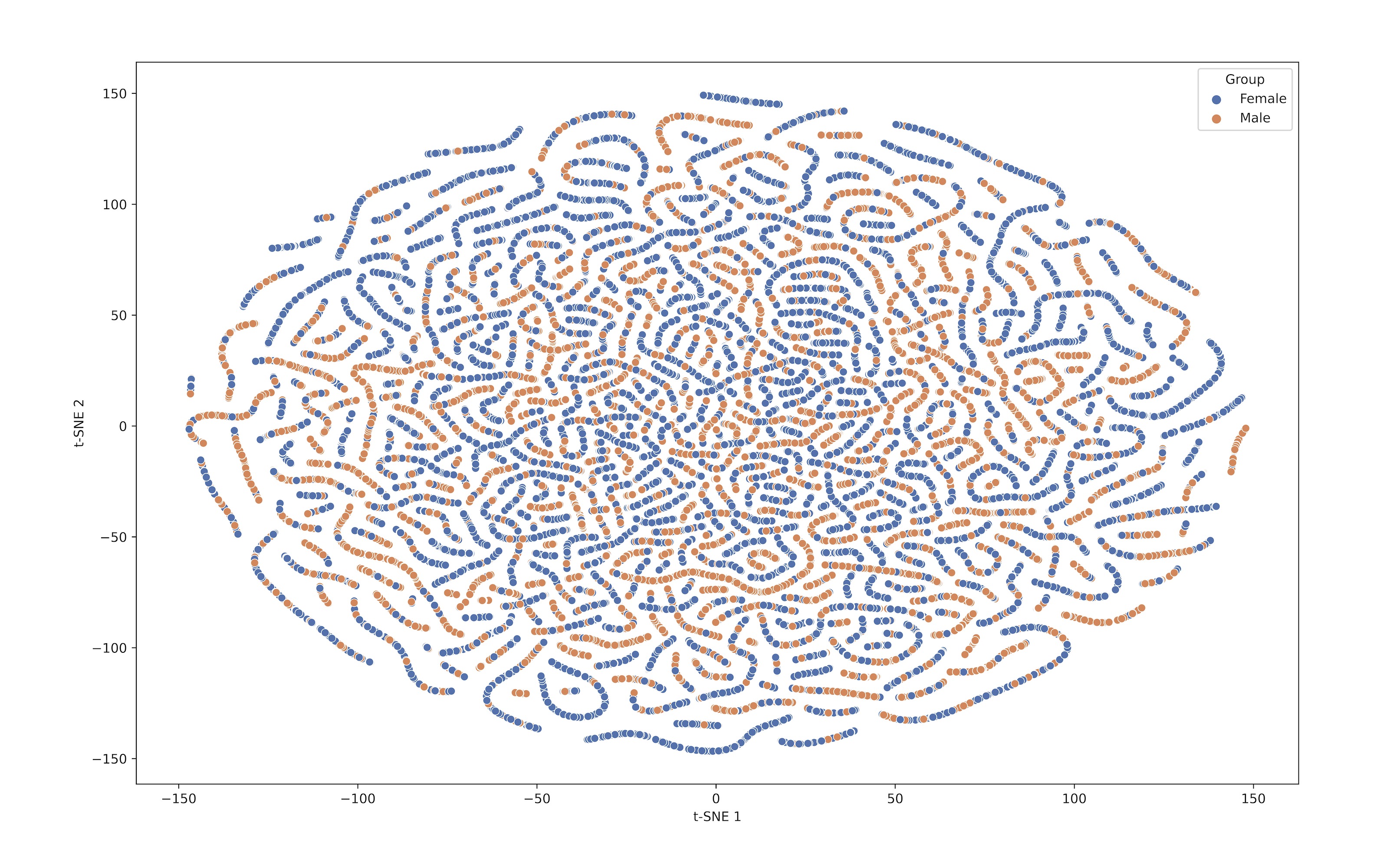}%
\label{fig:7}}
\caption{CelebA Dataset – t-SNE visualization $z$ without the Distraction module (a) and with the Distraction module (b) labeled with gender classes. The invariant encoding $z$ (a) shows clustering by gender. In contrast (b) shows no recognizable clustering. These plots are generated using "Attractive" attribute.}
\label{fig:tsne}
\end{figure*}
To further illustrate the effectiveness of the Distraction module, we present t-SNE visualizations of the output embeddings from the ResNet-18 model both with and without the Distraction module. t-SNE (t-distributed Stochastic Neighbor Embedding) is a dimensionality reduction technique that is particularly well-suited for visualizing high-dimensional data in a low-dimensional space.\\
In Figure \ref{fig:4}, we show the t-SNE plot of the embeddings, $z$, produced by the ResNet-18 model without the Distraction module. In this plot, it is evident that the embeddings are clustered by gender, suggesting that the model has learned to rely on gender information for its classifications. This clustering by gender is problematic as it indicates that the model is biased and may exhibit unfair behavior when making predictions.\\
Conversely, in Figure \ref{fig:7}, we present the t-SNE plot of the embeddings, $\tilde{z}$, generated by the ResNet-18 model with the Distraction module. In this visualization, the gender-based clustering observed in the previous plot is no longer apparent. Instead, the embeddings are distributed more evenly in the low-dimensional space, indicating that the Distraction module has successfully reduced the influence of gender information on the model's embeddings.\\
The comparison of these two t-SNE plots demonstrates the capability of the Distraction module to mitigate bias in the model's embeddings. By preventing the model from relying on protected attributes such as gender for its classifications, the Distraction module promotes fairness and reduces the risk of discriminatory behavior in the model's predictions.

\subsection{Ablation Study}
\label{ap:ablation}
 In this subsection, we perform an ablation study to investigate the effects of different functions for the fairness layer in the Distraction module. The fairness layer can be any differentiable function with controllable parameters denoted as $\theta_d$. We experimented with three configurations for the Distraction module: one linear layer, two linear layers, and three linear layers on tabular datasets. The results of the ablation study are summarized in Table \ref{tab:ablation_study}.
\\
For the CelebA dataset, we explored three types of fairness layers: linear layers, Residual Blocks (ResBlocks), and Convolutional Neural Network (CNN) layers. The mean scores of each category of CelebA attributes for each type of fairness layer are provided in Table \ref{tab:ablation_study2}. The "Inconsistency between experts" (Incons.), "Gender-dependent" (G-dep), and "Gender-independent" (G-indep) columns in Table \ref{tab:ablation_study2} represent different fairness metrics related to the CelebA dataset.
\\

The justification for the performance differences between the ResBlock and the fully connected models in our ablation study lies in the proportion of the model occupied by the Distraction module and the specific contributions of these modules to different parts of the network. In particular, there are two primary factors that explain the observed performance differences: the roles of the modules in the network and the flow of data through these modules.

\textbf{Role in the Network}: The ResBlock and the fully connected modules serve different purposes within the network. The ResBlock contributes to the embedding space of the image, which includes feature extraction and representation learning. This enables the model to capture the essential characteristics of the image while minimizing the effect of the protected attributes (e.g., gender) on the classification task. In contrast, the fully connected module is mainly involved in the classification part of the network, where it contributes to the decision-making process based on the features extracted from the previous layers. This distinction in roles explains why the ResBlock provides more fair results, as it directly affects the representation learning and reduces the influence of the protected attributes on the embeddings.\\
\textbf{Flow of Data}: The flow of data through the ResBlock is different from the flow through the fully connected and CNN modules. ResBlocks have skip connections that allow the input to bypass some layers and directly flow to the subsequent layers. These skip connections help in preserving the original information and preventing the loss of critical features during the network's forward pass. As a result, the ResBlock is more effective in capturing the inherent relationships in the data while mitigating the bias from the protected attributes \cite{resnet}. In contrast, CNNs involve multiple convolution and pooling operations, which can cause the loss of some information relevant to fairness. The fully connected module, with its dense layers, lacks the skip connections present in the ResBlock, which can lead to less effective bias mitigation.\\

In conclusion, our ablation study demonstrates that the choice of fairness layer in the Distraction module can significantly impact the fairness and accuracy of the model. It is essential to strike a balance between fairness and accuracy and to select the appropriate fairness layer for the specific dataset and application at hand.

\begin{table}[t]
	\caption{Area over the curve of statistical demographic parity and accuracy for model ablation}
	\label{tab:ablation_study}
	\begin{center}
		\begin{small}
			\begin{sc}
				\begin{tabular}{lcc}
					\hline
					Method & UCI Adult & Heritage Health \\
					\hline					
					One Linear Layer & \textbf{0.411} & 0.492 \\
					
					Two Linear Layers & 0.404 & 0.513 \\
					
					Three Linear Layers  & 0.349 & \textbf{0.531} \\
					
					\hline
				\end{tabular}
			\end{sc}
		\end{small}
	\end{center}

\end{table}

\begin{table*}
	\caption{Accumulative comparison between different Distraction layers}
	\label{tab:ablation_study2}
\centering
\begin{tabular}{c|ccc|ccc|ccc}
\hline
CNNBlock         & \multicolumn{3}{c}{AP}                               & \multicolumn{3}{c}{$\Delta$DP}                       & \multicolumn{3}{c}{$\Delta$EO}                       \\
\hline
                 & \textbf{Incons.} & \textbf{G-dep} & \textbf{G-indep} & \textbf{Incons.} & \textbf{G-dep} & \textbf{G-indep} & \textbf{Incons.} & \textbf{G-dep} & \textbf{G-indep} \\
\hline
One Linear Layer & \textbf{0.646}            & \textbf{0.755}          & \textbf{0.841}            & 0.072            & 0.115          & 0.085            & \textbf{0.084}            & 0.069          & 0.089            \\
CNN Res Block    & 0.568            & 0.699          & 0.768            & \textbf{0.04}             & \textbf{0.035}          & \textbf{0.026}            & 0.126            & \textbf{0.067}          & \textbf{0.062}            \\
CNN Layer        & 0.617            & 0.731          & 0.822            & 0.058            & 0.092          & 0.069            & 0.099           & \textbf{0.067}          & 0.073           
\end{tabular}
\end{table*}

\section{conclusion}
In this paper, we introduced a novel bias mitigation strategy, the Distraction module, for training deep learning models on various data types, including tabular, image, and graph data. Our proposed approach addresses the problem of unwanted bias in machine learning models, which can lead to algorithmic discrimination and unfairness towards certain population subgroups, particularly minority groups. Unlike existing methods that may result in information loss or fail to balance accuracy and fairness, the Distraction module effectively controls bias while maintaining model performance.

The Distraction module operates within the main neural network, independently managing a subset of weights and optimizing towards specific criteria or alternative objective functions. This in-process bias mitigation method does not require adversarial example generation and can be trained on available datasets without any alterations.

Our empirical evaluation on multiple benchmarks, including the UCI Adult, Heritage Health, POKEC-Z, POKEC-N, NBA, and CelebA datasets, demonstrates the effectiveness of our approach. Compared to state-of-the-art methods in the fairness literature for each dataset, our model exhibits superior performance in minimizing bias while preserving accuracy. We also introduced a new adversarial training procedure that further enhances both accuracy and fairness metrics.

In conclusion, the Distraction module provides a powerful and flexible strategy for mitigating bias in deep learning models. Its ability to operate within the main neural network and optimize specific criteria enables more precise control over bias without sacrificing model performance. As a result, the Distraction module offers a promising approach for addressing the ethical concerns associated with algorithmic bias and promoting fairness in artificial intelligence applications.

\section{acknowledgement}

\bibliographystyle{IEEEtran}
\bibliography{bib}


 





\end{document}